\documentclass[11pt]{article}
\usepackage[utf8]{inputenc}
\pdfoutput=1

\usepackage{booktabs} % For formal tables
\usepackage{fullpage}
\usepackage{xfrac}
\usepackage{algorithm}
\usepackage[noend]{algorithmic}
\usepackage{amsmath,amsthm,amsfonts}
\usepackage{amsmath}
\usepackage{amssymb}
\usepackage{color}
\usepackage{mathrsfs}
\usepackage{verbatim}
\usepackage{enumitem}
\usepackage{bm}
\usepackage[pagebackref]{hyperref}
\usepackage[capitalise,noabbrev]{cleveref}
\usepackage{xr}
\usepackage{tikz}
\usepackage{subcaption}
\usepackage{float}
\usepackage{listings}
\usepackage{array}
\usepackage{makecell}

\usepackage[suppress]{color-edits}
\usepackage[]{color-edits}
\addauthor{ll}{blue}

\definecolor{DarkBlue}{rgb}{0.1,0.1,0.5}
\hypersetup{
	colorlinks=true,       % false: boxed links; true: colored links
	linkcolor=DarkBlue,          % color of internal links
	citecolor=DarkBlue,        % color of links to bibliography
	filecolor=DarkBlue,      % color of file links
	urlcolor=DarkBlue,          % color of external links
	pdftitle={},
	pdfauthor={},
}

\usepackage{tabularx}
\usepackage{xcolor}
\usepackage{float,subcaption,placeins}
\usepackage{amsmath}
\usepackage{amsthm}
\usepackage{thm-restate}
\usepackage{float}
\usepackage{enumitem}
\usepackage{xr}
\usepackage{graphicx}
\graphicspath{{./figures/}}
\usepackage{thmtools}
\usepackage{url}
\usepackage{bm}
\usepackage[normalem]{ulem}
\usepackage{balance}
\usepackage{lipsum}
\usepackage{etoc}
\usepackage{comment}
\newcommand{\Comments}{1}
\newcommand{\mynote}[2]{\ifnum\Comments=1\textcolor{#1}{#2}\fi}

\usepackage[numbers,sort&compress]{natbib}

\renewcommand\cite[1]{\citep{#1}}

\usepackage[group-separator={,}]{siunitx}
\usepackage{pgfplots}

\pgfmathdeclarefunction{gauss}{2}{%
  \pgfmathparse{1/(#2*sqrt(2*pi))*exp(-((x-#1)^2)/(2*#2^2))}%
}

\usetikzlibrary{patterns}

\pgfplotsset{compat=1.10}
\usepgfplotslibrary{fillbetween}

\usepackage{xcolor}
\definecolor{babyblueeyes}{rgb}{0.63, 0.79, 0.95}
\definecolor{powderblue}{rgb}{0.69, 0.88, 0.9}

\usepackage{chngcntr}
\usepackage{apptools}
\title{Lost in Translation: Reimagining the Machine Learning Life Cycle in Education}

\author{Lydia T. Liu\footnotemark[3]~\footnotemark[6] \thanks{L.T.L. and S.W. contributed equally to this work as co-lead authors.} \quad \quad Serena Wang\footnotemark[3]~\footnotemark[1] \quad \quad  Tolani Britton\footnotemark[4]~\thanks{T.B. and R.A. contribtued equally to this work as co-principal investigators.}  \quad \quad Rediet Abebe\footnotemark[3]~\footnotemark[2]\\
	\\
	\footnotemark[3]~Department of Electrical Engineering and Computer Sciences\\
	University of California, Berkeley \\
			\footnotemark[4]~Graduate School of Education\\
	University of California, Berkeley \\
	\footnotemark[6]~Computing and Information Science\\
	Cornell University
}
\date{}

\begin{document}

\maketitle
\begin{abstract}
Machine learning (ML) techniques are increasingly prevalent in education, from their use in predicting student dropout, to assisting in university admissions, and facilitating the rise of MOOCs. Given the rapid growth of these novel uses, there is a pressing need to investigate how ML techniques support long-standing education principles and goals. In this work, we shed light on this complex landscape drawing on qualitative insights from interviews with education experts. These interviews comprise in-depth evaluations of ML for education (ML4Ed) papers published in preeminent applied ML conferences over the past decade. Our central research goal is to critically examine how the stated or implied education and societal objectives of these papers are aligned with the ML problems they tackle. That is, to what extent does the technical problem formulation, objectives, approach, and interpretation of results align with the education problem at hand. We find that a cross-disciplinary gap exists and is particularly salient in two parts of the ML life cycle: the formulation of an ML problem from education goals and the translation of predictions to interventions. We use these insights to propose an extended ML life cycle, which may also apply to the use of ML in other domains. Our work joins a growing number of meta-analytical studies across education and ML research, as well as critical analyses of the societal impact of ML. Specifically, it fills a gap between the prevailing technical understanding of machine learning and the perspective of education researchers working with students and in policy. 
\end{abstract}

\section{Introduction}

The widespread use of machine learning (ML) across domains has been controversial, with experts exposing concerns around data curation, relevance and appropriate use of ML techniques, and the potential for algorithms to create and amplify inequalities. While widespread public conversations around the use of ML are a more recent phenomenon, the computer science community has employed ML approaches widely in tasks such as recommendation systems \cite{bennett2007netflix} and speech and image recognition \cite{jelinek1997statistical,bottou1994comparison,shavlik1990readings}. More recently, numerous other disciplines have turned towards ML to increase efficiency and improve outcomes. For example, ML algorithms are seeing an increase in use across education. They have been deployed in a variety of ways both at the secondary and postsecondary levels, often with a stated goal of improving student performance. Some of the uses include predicting student dropout at the secondary level \cite{bird2021bringing,tamhane2014predicting}, evaluating applicants in college and graduate school admissions, and predicting persistence in massive open online courses (MOOCs) \cite{He2015,ramesh2014learning,conati2018ai}.

Extensive research currently explores the use of machine learning for social good \citep{abebe2020roles,kleinberg2016guide,shi20survey,cowls2021definition}, or ``ML4SG'' . Despite a surge in interest in understanding the societal impact of ML, this research faces two prevailing challenges. First, empirical evidence on the long-term effectiveness of ML4SG remains sparse \citep{cowls2021definition}, given the relative novelty of the area. Second, despite the application of ML4SG in consequential applications—such as education, environmental protection, and healthcare—inquiry into what ``social good'' entails in these contexts, and the extent to which ML4SG efforts contribute to the relevant social goals remains nascent \citep{floridi2020design}.

Despite the proliferation of ML4SG research in education, a number of recent algorithmic solutions in education \cite{waters2014grade, smith2020algorithmic, engler21} have led to negative and disparate outcomes, with students from historically marginalized backgrounds bearing the brunt of the burden. While these instances of harm highlight the importance of interdisciplinary collaboration to evaluate the ethics, equity, and impact of ML techniques in the education domain, for instance via the scrutiny of data sources \cite{yu2020towards}, surveillance practices \cite{lu2021data}, and transparency of the ML models \cite{bird2021bringing,conati2018ai}, they also demonstrate that a gap persists between the intent and impact in ML for education domain applications.
% Thus, we expand this discussion to evaluate the gaps between the research papers published applying ML to education and the actual impacts (intended or unintended) that these papers may have.
% The publication and adoption of ML research papers in education has significant ramifications for stakeholders, and we thus focus on the contributions that ML researchers can make in more conscientiously aligning the research objectives, ML problem formulation, and discussion of interventions with actual intended outcomes.

% The rapid proliferation of ML driven technologies in the education sphere means it is more urgent than ever to address their societal impact to bridge the divide between intent and impact.

Multidisciplinary research communities in education, including educational data mining (EDM)  \cite{labarthe2018analyzing, calders2012introduction}, learning analytics (LA), Computer-Supported Collaborative Learning (CSCL) \cite{lin2018examining}, and AI for education (AIED) \cite{chen2022two, pinkwart2016another} have been active, with overlapping research interests and philosophical differences described in \cite{rienties2020defining}.
Surveys of research trends and challenges in these communities \citep{guan2020artificial, chen2022two, holmes2021ethics} have indicated rapid growth in the use of technology in education and consequent research interest. This growth necessitates a deep dive into the technological priorities in applied ML research and development, which is central to the current work.  A shared interest in common problem formulations and technologies notwithstanding, Calders et al. \cite{calders2012introduction} suggest that there are salient disciplinary differences between education research communities and traditional ML and data science communities such as KDD\footnote{ACM SIGKDD International Conference on Knowledge Discovery and Data Mining}  \cite{piatetsky2012introduction}.
% For example, relatively few EDM researchers came from a ``data mining background''.

%Works like Chen et al. \cite{chen2022two} and Guan et al. \cite{guan2020artificial} focus on overall research trends in interdisciplinary publication venues, asking questions like ``What were the most investigated research topics?'' \cite{chen2022two}.

Our work focuses on the \emph{cross-disciplinary gaps} found within mainstream ML communities, and in particular, the gaps between persons developing new ML methodologies for education within a ML community and education experts who research and evaluate the use of technologies in schools and classrooms. We ask,  \textit{``How does the technical setup of ML papers (objective, evaluation metrics, modeling techniques) match educational goals and principles?''} Prior work conceptualizes ``epistemic trespassing'' \cite{watkins2022four}, where algorithmic contributions from technical researchers overlook important applied context and critical perspective. We specifically study how such gaps surface, through the expertise of education researchers and practitioners, as machine learning researchers formulate prediction tasks in highly complex education settings.

Many research landscape studies \cite{chen2022two, guan2020artificial} focus on overall research trends in the aforementioned education publication venues. Building on this research, we examine cross-disciplinary research practices from the perspective of ML technology development, and focus on contributions to the education domain from broad machine learning and AI conferences such as KDD, AAAI\footnote{The AAAI Conference on Artificial Intelligence}, and IJCAI\footnote{International Joint Conference on Artificial Intelligence}, over the past decade. These have an audience of computer and data scientists whereas education technology conferences tend ``towards smaller, more focused conferences and communities'' with primarily education researchers \cite{pinkwart2016another}. While ML and AI conferences have published highly cited ML4SG research relevant to education \citep[such as][]{waters2014grade, lakkaraju2015machine}, these articles are not typically included in survey studies \cite{chen2022two} published in multidisciplinary education technology venues.

%\llcomment{Condense above paragraphs to 1) introduce the edtech community and how they overlap with ML communities. 2) Retrospective work in AIED has grappled with related/overlapping challenges, such as validation and human operators, but not the technical ML process, Also they do not include the ML conferences with significant ml4Ed research. }

We carry out in-depth semi-structured interviews with 15 education researchers and practitioners to critically examine the divide between intent and impact in the existing research literature of ML4SG applied to education (``ML4Ed''). Interviewees have domain backgrounds in higher education policy and K-12 education policy. We include researchers across  K-12 and higher education because many of the challenges with ML technology use occur in both settings. The interviews were centered around a discussion of research papers on ``AI for social good'', compiled by Shi et al. \cite{shi20survey}, that are relevant to ML4Ed and represent recent scholarship appearing in mainstream machine learning and AI conferences.
Specifically, we explore the formulation of the machine learning task, the role and function of prediction, and whether the intended and realized impacts on students are well-aligned.

 Prior critical work explored cross-domain literature surveys, such as surveys of AI4SG \cite{floridi2020design, shi20survey} and of algorithmic harms and bias \cite{suresh2019framework,blodgett2020language}. We focus our investigation on a single application domain---education---and draw on insights from  interviews with education experts to answer our research questions on the gap between machine learning \emph{research} practices and the use of ML technologies in education. In contrast, prior fieldwork-based studies explored the practice of machine learning in industry settings \cite{passi19problem, haakman2021ai}. By engaging education researchers in discussions of current work in ML4Ed, we use a  cross-disciplinary lens to bring to light facets traditionally overlooked in ML research. %Education researchers are uniquely positioned to evaluate academic papers which are our artifacts of study, and many possess on-the-ground experience applying data science and ML methods to education domain problems.
While study participants do not directly speak for the lived experiences of stakeholders such as students, teachers, parents and institutions, we draw upon their nuanced and broad understanding of various stakeholders' perspectives and interactions with these stakeholders.

While we focus on education as a specific and important domain, many of our findings can extend to other domains and highlight common areas for improvement in machine learning practices. By studying ML communities' forays into education applications and teasing out both education-specific lessons and general lessons that apply to applications of ML in other domains, our work answers  Holmes et al.\cite{holmes2021ethics}'s call to %the AIED community to 
increase the ``impact of AIED research on the increasingly human-oriented, real-world applications of AI'', but focusing on ML4Ed in broad ML/AI conferences. %Even within the AIED communities, there is concern about cross-pollination with broader AI research communities, as Holmes et al. \cite{holmes2021ethics} call for increasing the ``impact of AIED research on the increasingly human-oriented, real-world applications of AI.'' In a sense, our work answers this call by studying AI communities' forays into education applications and teasing out both education-specific lessons and general lessons that apply to applications of AI in other domains.

 %not captured by existing frameworks.
We present two main findings in (Sections~\ref{sec:problem_formulation} and \ref{sec:interventions}), which correspond to the misalignment of machine learning experts and education researchers with respect to the problem formulation and the limits of prediction tasks. For the first finding, narrowing multidimensional outcomes to a single quantifiable metric that can be computed by an algorithm can lead to oversimplification of complex educational problems and the neglect of education equity and access goals. This theme was echoed by numerous educational researchers and practitioners. Our second finding highlights that prediction is useful when it provides actionable information for interventions to improve outcomes, but can lead to negative outcomes if prediction is treated as the primary goal. Given these findings, we discuss improvements to machine learning systems and approaches used in education to increase the likelihood of positive educational outcomes for students. In Section~\ref{sec:dis}, we discuss the connections to relevant multidisciplinary literature including education technology, human-computer interaction (HCI), and fairness, accountability, transparency, and ethics (FATE) in computer science.

% Example PNAS ed & qual paper: https://www.pnas.org/content/118/37/e2026386118
% Example PNAS survey-based paper: https://www.pnas.org/content/117/31/18317'

\section{Data and Methods}

This study is based on data generated from in-depth semi-structured interviews with 15 education researchers discussing selected research papers that apply ML to education (``ML4Ed''). 

% Guiding questions for Method

% \begin{itemize}
%     \item  \emph{what is motivating your study and why your methods are appropriate. For inductive studies, articulating one's motivation not only involves reviewing the literature to illustrate some “gap” in prior research, but also explaining why it is important to fill this gap.}
%     \item \emph{Are you building new theory or elaborating existing theory?}
    
%     New theory: ML life cycle in education. The idea of an ML life cycle exists informally in the ML and Data Science community (also in a Computer systems sense), but does it exist in a formal/academic setting, and (how) do we want to engage with that notion?
    
%     Existing theory on data science life cycles, e.g. CRISP-DM?
    
%     \item \emph{Why did you choose this context and this “unit of analysis?” Qualitative methods paper writers should explain the nature of the context they are examining. Similarly, qualitative authors should discuss whether they are sampling people, events, cases, and the like, and why they are being sampled.}
% \end{itemize}

    \paragraph{ML4Ed papers}
    
  % Prior to data collection, we first performed two separate sampling procedures---of ML4Ed research papers and of education researchers to recruit as interviewees.
    
    The papers we discussed during the interviews were sampled from a dataset of research papers on ``AI for social good'' compiled by Shi et al. \cite{shi20survey}. The dataset contains 1176 papers published between 2008 and 2019, of which 78 papers were labeled as ``education''-related by Shi et al. via keyword matching. We then selected papers from this set that were relevant to machine learning based on their abstracts\footnote{We initially filtered the 78 education-related papers by the presence of keywords (``predict'' and ``machine learning'') in their abstracts. We then finely selected the papers by reading the abstracts of each paper to ensure that the selected paper involved (1) formulating and solving a machine learning problem and (2) an education-related application. The last step was performed using the authors' respective expertise in machine learning (L.T.L., S.W. and R.A.) and in education (T.B.).}, leading to a list of 20 papers (``ML4Ed list'') \cite{lan2014time,huang2017question,yin2019quesnet,vie2019knowledge,mi2015probabilistic,ramesh2014learning,He2015,gupta2019deep,wu2019zero,lakkaraju2015machine,waters2014grade,xu2017progressive,tamhane2014predicting,shashidhar2015spoken,grover2019semantics,woods2017formative,labutov2017semi,roy2019inferring,nosakhare2019probabilistic,su2018exercise}. We randomly examined unselected papers and confirmed that there was no significant exclusion of papers pertinent to ML4Ed by our sampling process. Major topics in education that are covered by the final set of papers include higher education, student learning, MOOCs, and standardized assessment. Notable omissions include special education, early education, and teaching.\footnote{The education topics were identified by T.B., who works in education research, based on paper titles and abstracts.}

      \paragraph{Participants}
    Interview participants were recruited through purposive sampling \cite{patton2002qualitative, palinkas2015purposeful} from the authors' professional networks %(see author positionality statement at the end of this section). 
    In order to be invited for our study, participants needed expertise in education research and practice. The current roles of the participants include Ph.D. candidates in education and economics of education, postdoctoral researchers,  university faculty members, and research directors at public and private education agencies. In terms of education level, we included a mix of participants with expertise in K-12 and higher education. We refer to them anonymously as P01-P15. Table \ref{table:participants-summary} lists all participants' self-reported occupations, research areas, genders, and races/ethnicities. In addition to current occupations, Table \ref{table:experiences} lists other experiences in the education sector that participants have had. Further details that participants provided on their backgrounds indicated common experiences with teaching in public high schools, engagement in policy evaluation, and non-profit work focused on students from working class families.
    
    \begin{table*}[!ht]
    \centering
    \resizebox{\textwidth}{!}{
    \begin{tabular}{|l|l|l|l|l|}
    \hline
        \textbf{Participant} & \textbf{Occupation} & \textbf{Research area keyword} & \textbf{Gender(s)} & \textbf{Race(s)/ethnicity(ies)} \\ \hline
        P01 & Executive director and full professor & tertiary student success data analytics & M & White \\ \hline
        P02 & Director of research & economics of education & F & Asian \\ \hline
        P03 & Associate professor & higher education & F & White \\ \hline
        P04 & Senior lecturer & education policy & F & White \\ \hline
        P05 & PhD candidate & college admissions & M & Hispanic and Latino \\ \hline
        P06 & PhD candidate & higher education & (not reported) & (not reported) \\ \hline
        P07 & Vice president & higher education & F & White \\ \hline
        P08 & PhD student & K-12  & M & African American \\ \hline
        P09 & Assistant professor & K-12 & M & Asian \\ \hline
        P10 & Senior director & tertiary student success & M & White \\ \hline
        P11 & Associate professor & K-12  & F & White, MENA/SWANA \\ \hline
        P12 & Assistant professor & learning at scale & M & White \\ \hline
        P13 & Full professor & MOOCs & M & Asian \\ \hline
        P14 & Postdoctoral researcher & education policy & F & Asian, White \\ \hline
        P15 & PhD candidate & economics of education & F & Hispanic and Latino \\ \hline
    \end{tabular}
    }
    \caption{Summary of study participants. Gender, race, and ethnicity information are voluntarily self-reported.}\label{table:participants-summary}
    \end{table*}
    
    \begin{table}[!ht]
    \centering
    \begin{tabular}{|l|c|}
    \hline
    \textbf{Education sector experience} & \textbf{\# of participants} \\ \hline
    Taught at a university & 10 \\ \hline
    Created or designed curriculum & 9 \\ \hline
    Built tools used in the education sector & 8 \\ \hline
    Taught at an elementary/middle/high school & 7 \\ \hline
    Worked with policymakers & 7 \\ \hline
    Worked with not-for-profit organization(s) or NGO & 6 \\ \hline
    Worked with for-profit organizations in education & 5 \\ \hline
    \end{tabular}
    \caption{Education sector experiences that participants have had aside from their listed occupations in Table \ref{table:participants-summary}.}\label{table:experiences}
    \end{table}

  \paragraph{Matching} Each interview focused on one paper from the aforementioned ML4Ed list. We matched papers to participants by respecting the participant's preferences. Participants filled out a pre-interview survey (\href{http://}{SI Dataset S1}), indicating up to six papers that they are ``willing to discuss'' or that they ``prefer to discuss'' during the interview. 
 The authors selected one of the participant's preferred papers for discussion during the interview, prioritizing the coverage of education topics %(see Table~\ref{table:topics-summary-20} in Appendix~\ref{app:supp_methods} 
 (see \href{http://}{SI Dataset S2, Table S1, Table S2} for the list of papers and topics) whenever possible. Finally, nine unique papers were discussed during the interviews.

%     Comment on role of papers discussed.
% A robust interview process - grounded by paper.

    \paragraph{Interviews}
    
    Our interview data were generated from December 2020 to September 2021. Interviews lasted between 50 to 60 minutes, and were conducted through Zoom.  Participants were sent the relevant ML4Ed paper beforehand for voluntary perusal. We asked participants' permission to record the interview via a consent form, as well as verbally confirmed their consent before beginning to record the interview. Each interview comprised of two parts: 1) introductions and background questions 2) a focused discussion on the research paper. Part 1 included questions on the participant's current research interests in education, and on the impact of data science or machine learning in education contexts that they are familiar with. After contextualizing the participant's professional experience, we began Part 2 by reading together selected sections of the ML4Ed paper, and followed the reading of each section with discussion prompts such as ``How would you describe their goals?'', and  ``To what extent do you feel this machine learning task captures the ... goal?''. All interviews were recorded and transcribed.

% If good: 
% Why do you think it's good? What metrics by researchers in education and are they aligned? 
% Would you anticipate heterogeneous effects from this proposal?
% If neutral/bad: 
% What does it seem to be capturing well/ok? 
% What is it missing? 
% Do you have suggestions for proposals that would mitigate this? 
% Do you think there could be student groups harmed? If so, which student groups are helped/harmed by this approach?
% Findings: 
% To what extent do these mirror what you'd expected? 
% Is this in line with findings in the education literature? 
% How do think these findings could be used in education? 

%... \todo{what the interviews were like, how did the papers ground the discussion, what kinds of question did we ask, how structured was it?}
    \paragraph{Data Analysis}
    We analyzed the data inductively. In the initial stage, all four authors conducted open coding on Atlas.TI and met biweekly to compare generated codes and discuss emergent themes. From the codes generated by individual authors, a preliminary list of codes were identified through consensus that included themes such as \emph{education goals of machine learning technologies}, \emph{from machine learning tasks to interventions}, \emph{problem complexity} and \emph{human stakeholder engagement}. The authors returned to code selected interviews with the preliminary list of codes, individually, and in subsequent meetings discussed and reconciled differences, and continued to refine the salient themes. This process was iterated until the authors reached consensus on the final list of themes and how they were assigned to text. Finally, the first two authors labeled each theme with a short name, wrote up informative definitions, and re-coded the dataset in its entirety. 
    
   % \paragraph{Author positionality} The first two authors are graduate students in Computer Science; the third and fourth authors are faculty members in Education and Computer Science respectively, all in the United States. All authors identify as women.\\
    
    Further details on our data and methods, including summaries of participant information (\href{http://}{SI Dataset S3}), interview questions (\href{http://}{SI Dataset S4}), and codes (\href{http://}{SI Dataset S5}), can be found in the supporting material. %Appendix~\ref{app:supp_methods}.

\section*{An Extended ML Life Cycle}

\begin{figure*}[tbp]
    \centering
    \includegraphics[width=\textwidth, trim = 20 150 20 120, clip]{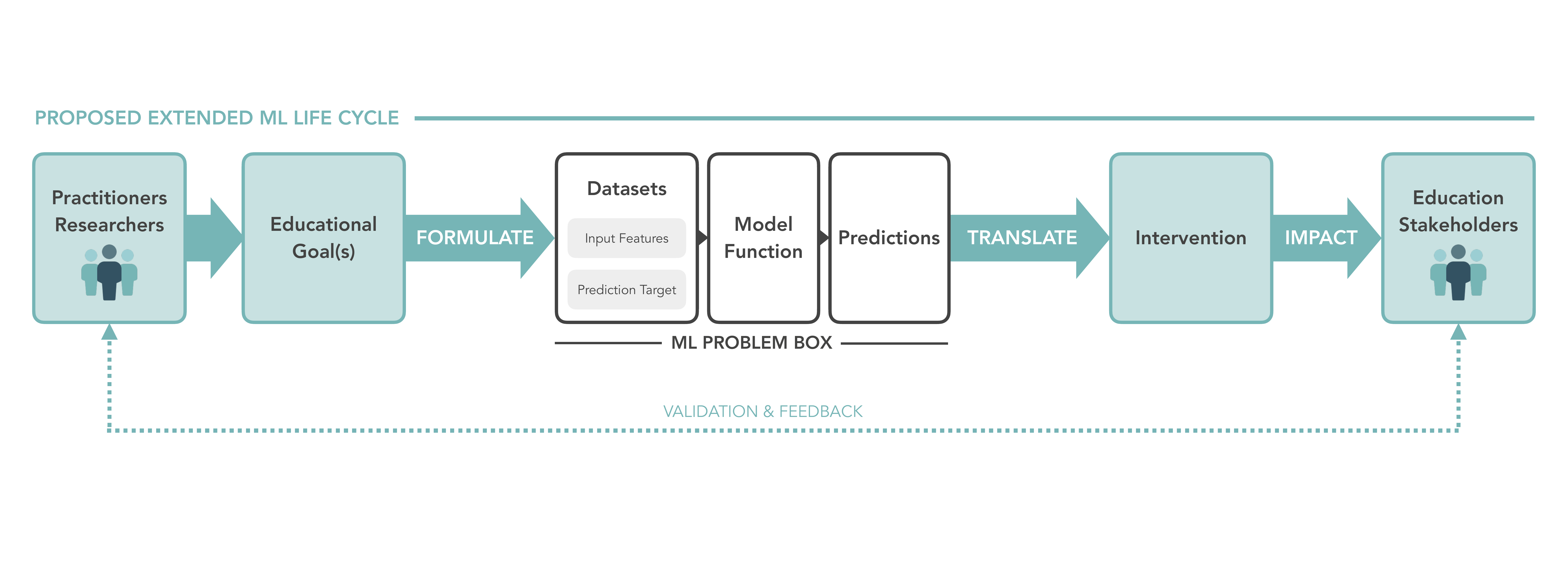}
    \caption{An Extended ML Life Cycle Diagram. The inner ``ML Problem Box'' represents the typical aspects of the ML problem detailed in the surveyed ML research papers. Our interview findings reveal the need to consider an extended version of the ML life cycle in ML research, including the initial problem formulation stage by practitioners and researchers, and the translation from predictions to interventions that eventually impact stakeholders.}
    \label{fig:process}
\end{figure*}

To frame our findings, we start by providing an overview of the ML life cycle as it pertains to the education papers surveyed. We illustrate this in the Extended ML Life Cycle Diagram (Figure~\ref{fig:process}), and refer to this as a conceptual tool to capture the key findings from our interview study.
We also define the terminology we use for the different parts of the ML process.

Grounding this terminology in the education domain, we consider a running example of predicting student dropout risk, which was a common application among the papers from Shi et al.\cite{shi20survey}'s Survey on AI for Social Good  \cite{He2015,lakkaraju2015machine,xu2017progressive,tamhane2014predicting}. In a typical application of supervised machine learning, the practitioner decides on a particular \textit{dataset} to use for model training and validation (e.g. historical student data over the past several years). The dataset comprises \textit{input features}, denoted $X$ (e.g. student grades, attendance, demographic information), and a \textit{prediction target}, denoted $Y$ (e.g. whether or not a student dropped out of a program). The goal of the practitioner is to find a \textit{model function} $f$, typically through mathematical optimization, such that the output of $f$ when applied to the input features, $f(X)$, closely approximates~$Y$, not only on the dataset but also on new samples. This procedure is referred to as \textit{training}. $f(X)$ is said to output \emph{predictions} for~$Y$. ML papers often focus on the family of functions over which $f$ is trained, such as regularized linear models, neural networks, and decision trees.

The aforementioned elements make up the \textit{ML Problem} box pictured in Figure~\ref{fig:process}. In the ML4Ed papers surveyed, these elements are typically prominently featured and discussed. Indeed, these elements can have significant ethical implications and impact on outcomes. For instance, sampling and historical bias in the input features and prediction target have been scrutinized in the literature on fairness, accountability, transparency, and ethics (FATE) \cite{suresh2019framework,Obermeyer19dissecting}. 

However, as much as scrutiny of the ML Problem is crucial, we found that much of the critical discussion on ML4Ed research from our interviews fall \emph{outside} the traditional ML problem box -- in fact, all fifteen interview participants raised discussion points outside of the traditional ML problem box in the following ways. First, before the ML problem is even \textit{formulated}, practitioners and researchers must identify a set of \emph{education goals} to address (in the running dropout risk prediction example, these goals may include increasing on-time graduation rates). This influences the choice of input features, prediction target, and success metrics used to evaluate the model function. Once the model function is trained,  the subsequent \textit{interventions} that apply the \textit{predictions} produced by the trained model function are frequently overlooked in the ML research papers surveyed (in the risk prediction example, \emph{who} are the dropout risk scores shown to?). Thus, Figure \ref{fig:process} extends the ML life cycle outside of the ML problem box to include these consequential steps. Mirroring these important aspects of the ML life cycle that may be overlooked by ML4Ed research papers, our research findings are organized around two translational challenges:

\begin{enumerate}
    \item \textbf{Translating Education Goals to ML Problems.} How education goals are interpreted by computer science researchers and translated into a machine learning problem; 
    \item \textbf{Translating Predictions to Interventions.} How the predictions of ML models are `translated' into interventions, and how to evaluate and address their impact.
\end{enumerate}

\noindent In the following sections, we elaborate on the above findings from our interview study.

% Due to its nature, the issue of structural and resource inequality permeates our discussion of both. \todo{tbd?}

\section{Translating Education Goals to ML Problems}\label{sec:problem_formulation}
We first discuss issues and recommendations that arise in the process of translating education goals into the ML problems detailed in the ML4Ed papers. Many impactful design decisions arise just in this problem formulation stage, including in the choices of success metrics, targets, and inputs. 

%\subsection{Selective Interpretation of Education Goals}
\subsection{Conflating education goals with quantitative, short-term metrics}

While ML4Ed research papers often sought to address education goals, such as student learning and student success, they tended to reflect a narrow understanding of these goals by focusing exclusively on a single quantitative metric. Ten (out of fifteen) interview participants identified limitations in the single quantitative metric highlighted in ML4Ed papers spanning multiple problem contexts, from \emph{secondary success} and \emph{essay scoring} to \emph{college persistence} and \emph{graduate admissions}. For example, an ML4Ed paper may choose to tackle a specific goal related to a single metric such as ``improving on-time graduation rates''. Interviewees pointed out that though the chosen metric, graduation rate, is relevant in the context of student success, there are other metrics that also should be taken into account in order to capture student success (five participants\footnote{P01, P02, P03, P08, P09.}). %\lnote{How to indicate, and track the data for, how often an opinion was expressed?}

\begin{quote}
    P01 (Executive Director and Professor):
    ``We don't limit ourselves to graduation rates. We're also looking at other things such as retention rates and so forth.''
\end{quote}

\begin{quote}
    P02
    (Director of Research):
    ``Graduating with satisfaction or knowing exactly what they want to do afterwards. That's another thing that we could look at.''
\end{quote}

\noindent While both P01 and P02 agree that graduation rate is a useful metric, they each mention other important metrics that are possibly longer term or more difficult to quantify. 

%\todo{Keep or remove direct quotes?}

\paragraph{Lack of substantive justification}
While the single metric might be an important outcome, as in the case of college graduation rates, substantive justification for why  metrics matter often seems to be missing, beyond  correlation with longer-term student outcomes or the existence of data for the metric. 
P03, a faculty member in a school of education observed that in ML4Ed research ``there is often bias towards shorter term outcomes
without drawing out the logical map of why do we care'' because ``there is better data about them [...] they're more often in the same dataset''. Accounts of why the chosen metric is important may be overly inflated and more importantly, inaccurate to educational realities. 

% \begin{quote}
%      P03 (GSE Faculty): ``I'm just like not convinced that there is this causal mechanism that because I graduated a semester late, I am therefore going to have worse outcomes.''
% \end{quote}

% central theme: single metric is important doesn't capture rich tapestry of student experience. start the section. graduation or exam score. not always clear why we focus on metrics. -> what it means about 

% single scores have been used for multiple purposes from student success to graduate admissions. then refer back to the each example.

% "narrow interpretation of education goals..." is an important para.

 The focus on one metric also leads to overemphasis on exam scores as the learning metric. Commenting on the some ML4Ed research papers' usage of exam scores as a target, P04 (Senior lecturer) observes, ``It just sounds like somebody who doesn't actually know schools. It sounds like somebody who just sort of like has a theory in mind of how tests matter% and they're just trying to justify their research question
.'' P04 discusses how persons might focus on standardized exam scores without a deep understanding of what they measure and how they are related with student outcomes. 

Another example of this focus on exam scores comes from automated scoring of assessments. In the context of student learning, the focus on automated scoring centers the role of performance on assessments, as opposed to other aspects of learning, such as self-expression and creativity. For example, P05, a PhD candidate in education, finds that an automated essay scoring tool ``assumes that there is a universal good way to write''. While automated feedback may be useful to a student, to ``help point little things out'', any such tool necessarily encodes ``a lot of values [...] that [are] not being made transparent'' (P05, PhD candidate).
Moreover, the possibility of automating certain forms of assessment through technology warrants serious reflection on the relevance of such assessments, and whether they are meaningful for students.
\begin{quote}
    P05 (PhD candidate): ``If the writing is so mechanical that it's very easy for a computer to grade it like a human, then what are we even asking students to do in the first place?''
\end{quote}   The value of assessment as an education goal should not be taken for granted, as P05 notes above.

\paragraph{Competing goals and stakeholders}
In the context of graduate admissions, participants point out that education institutions can have competing goals. However, ML4Ed research may address one of them without acknowledging the tensions with the others. Driven by ``the current business model of higher education'', universities in the United States tend to optimize for ``efficiency'', according to P06, a PhD candidate in education. P06 further elaborates, ``they want students who would have higher rates of retention or graduation [...] otherwise, that's a financial loss to them''. However, many, especially public institutions, also have a mandate to promote education equity and access. Focusing almost entirely on making the admissions process ``more efficient'', the ML4Ed research paper proposed to use only past admissions decisions 
%what does it mean to only use past decisions?
to inform their applicant-rating algorithm while neglecting to address the concerns with education equity and access. In fact, if past application processes had led to lower rates of admissions for candidates with diverse academic profiles, an algorithm using this biased decision data likely led to similar admission outcomes.

Other than the institution's equity goal, the needs of key stakeholders, such as students, are also systematically left out by the selective attention that ML4Ed gives to particular education goals, as quantified by a single or a limited number of metrics, such as time savings. When asked whether a published admissions algorithm would meet educational needs, P07 (Vice president) responded, 
\begin{quote}
    P07 (Vice president): ``Whose needs? The student's needs, probably not. If you ask a student, do you want your application that you spend a lot of time on [...] to just go through an algorithm? I think they would say no. If you ask [...] your chief diversity officer, in their opinion, does this work well? They probably want to see, well, who's being admitted and who's being not admitted [...] For the faculty, yeah, it's working well because what they want is to spend less time and get high quality students admitted.''
\end{quote}
In this example, the interests of faculty do not converge with either the desire of students to have holistic review of their applications or the institutional interest in having a more diverse set of strong candidates for admission. 

%The narrow interpretation of education goals in ML4Ed research is not a mere artifact of the disciplinary disconnect between computer scientists and the education problems that they hope to address. What is concerning are the real and substantial harms that may arise from narrowly motivated ML4Ed research, should they make their way into policy discussions. P08 (PhD candidate) is concerned, for example, with the framing of education goals in ``pecuniary terms'' that neglect the ``non-pecuniary benefits to education'', suggesting that the chosen framing ``can lead to [...] perverse outcomes [...] beyond machine learning, beyond this paper''. 

%P05 (PhD candidate) expressed a related concern that an unquestioning attitude towards predicting  quantitative metrics with ML is having ``a negative impact'' on education and education policy more broadly. For instance, the participant suggests that the rise of automated scoring in various education contexts ``increases the belief that there's a need to quantify, to data-fy students that I don't think we actually need''. P05 opines that the introduction of more quantitative metrics is not necessary.

Although a single quantitative metric can contain important information, it rarely captures the rich tapestry of student experience, or the varied institutional objectives. Instead, the choice to focus on one or few metrics reflects ML4Ed's unexamined preferences for short-term and easily quantifiable education goals. In the following section, we delve deeper into the assumptions behind the choice of the single metric as the prediction target.

\subsection{Consequential choices in problem formulation: target and input} In the formulation of a supervised machine learning problem\footnote{Supervised learning was most common in the ML4Ed papers discussed.}, the choice of a prediction target and input features is paramount. Yet the value-laden nature of these choices is rarely examined in ML4Ed research.  Discussions with study participants suggest that expedient choices in problem formulation can have unintended implications for the research project and its applications down the road.

\paragraph{Universalizing individual narratives with prediction target}
In the education context, any choice of prediction target will inevitably lead to a loss of important information---the qualitative aspects of individual experience and narratives. Data that cannot be quantified in the prediction target is effectively dropped by a machine learning model. The irreducibility of qualitative perspectives is particularly salient in the automated evaluation of student writing. %P05
P05 (PhD candidate) worries about outweighing these with a ``zoomed out computational perspective''.

\begin{quote}
    %P05
    P05 (PhD candidate):  
    % ``For college admissions essays, I'm actually terrified that blunt force algorithmic solutionism is going to make its way into college admissions offices, because [...] 
    ``There's a lot of sociolinguistic variation in these essays. %And of course there is, obviously there is. 
     [...] You're asking students to write about themselves and [...]
     we have these unique, interesting experiences. And of course, they're going to be reflected in the essays.'' %[...] if you zoom out, with zoomed out computational perspective, or you zoom in and qualitatively code them.
\end{quote}

\noindent Similarly, in the context of graduate admissions, P06 (PhD candidate) stressed the importance of reconstructing individual narratives, as opposed to relying on quantitative summaries.
\begin{quote}
    P06 (PhD candidate): ``I think holistic admission is not like putting all these characteristics into a prediction model and then see[ing] what's the probability. [...] %it's more like 
    You're taking it into account these different pieces of information [...] to restore the actual life experience [...] difficulties they overcome throughout their life experience to achieve, to arrive where they are.''
\end{quote}

\noindent Both participants express reservations that ``blunt force'' (P05) algorithmic solutions may not adequately summarize the variation and unique narratives that are central to the holistic evaluation of student essays.

 In these settings, participants noted that ML4Ed researchers picked a single prediction target and evaluated individuals accordingly, thus demonstrating a ``universal[ist]'' attitude. According to P05 (PhD candidate), not only is ranking ``the best essays'' a misdirected way to think about the students' life stories, such an approach also assumes that the same metric applies to everyone. 
\begin{quote}
      P05 (PhD candidate):  ``Clearly the ideas of what's good writing in English are not transportable [...] in general, that's not a universal thing and these kinds of platforms, I think they're kind of built and designed with universality in mind, [that] any student can use this.%, but it's not that simple.
      ''
\end{quote}

\noindent P05 noted that formulating a machine learning problem with universality in mind runs the risk of discounting the rich and nuanced experiences of low-income students and students of color in particular, who may be underrepresented in the data.

%It is important to note that the problem is not the use of machine learning algorithms. Instead, the issue lies with the use of a single metric or ranking system based on machine learning technologies.  
\paragraph{Alternative problem formulations}
Problem formulations that do not rely on a single prediction target may be better aligned with the needs of the stakeholders. In the setting of graduate admissions, P06 (PhD candidate) questioned if instead of giving ``an explicit ranking''---based on ``whether a student, given their characteristics, would [have been]  admitted or not in the past''---the algorithmic system could  ``give summary information to the officers [...] to reduce the workload or [...] cognitive load'', which is ``even more aligned with people's initial demand.'' P06 suggests using the ML technologies to provide multiple dimensions of information on applicants. In this way, they avoid using biased past application data whilst potentially providing information on applicant strengths and areas of growth. 

\paragraph{Alignment of prediction target with education goals}
Even in cases where a single prediction target is appropriate, researchers might not end up selecting one that is well-aligned with education goals.  When predicting test outcomes, for example, the choice of test is important. One ML4Ed paper that sought to identify secondary students ``at risk'' of ``poor academic performance'' based its prediction target on student performance in two exams: a state test and a national standardized test.
\begin{quote}
   P04: ``[A general standardized test] more just measures like the opportunities they've had in their life in general as opposed to the curriculum they were supposedly getting in school [...] you were saying like they had more opportunity or more privilege in their background than actually measuring if they studied the material or if they learned.''
\end{quote}
  Here, P04 (Senior lecturer) questioned whether these tests directly measured a student's mastery of the specific content of their secondary curriculum, and if not, whether what they were measuring fit with the project's education goals.

 Another common practice in problem formulation is to convert a measurement, such as a test score, to a binary prediction target using an arbitrary threshold. In the context of identifying students who are academically at risk, this conflates students who are struggling academically with students who are just below the cutoff. P08 (PhD candidate) stressed that ``those are different groups'', alluding to the  `bubble kids phenomenon' in educational testing \citep{booher2005below}. Other participants expressed concern that this focuses attention unnecessarily on passing a particular assessment, however arbitrarily this may be defined.
 
 \begin{quote}
    P04 (Senior lecturer): ``Maybe the assessment is one piece of evidence among many that you might use to make that decision. But this person or the authors are making it sound like if you don't pass the test, you don't progress.''
 \end{quote}
Given the often arbitrary thresholds, basing decisions on how students will progress or the interventions that they might receive on prediction targets with arbitrary thresholds could lead to more negative outcomes for students. 

\paragraph{Expedient input features}
Apart from the choice of a prediction target, the choice of input data features is also significant. In several ML4Ed projects, participants including P12 (Assistant professor) found that this choice is dominated by convenience and data availability considerations. \begin{quote}
    P12 (Assistant professor): 
    %The other important thing about these features is it's
    ``[The input features are] just the stuff that we happen to be able to look at. It's not clear at all that this is what you would really want to know about a human being to figure out whether or not they were having a meaningful learning experience in a course.''
\end{quote}
Commenting on student data collected from MOOCs, P12 opined that ``these systems are not designed from the beginning to instrument things,'' and the data features that are easily available do not necessarily capture meaningful properties.

This inadvertently leads to a narrow and sometimes, overly individualistic view of student experience that downplays social and structural issues.
% \begin{quote}
%     P02: ``The big picture that we're missing [...] is that we're really focused on looking at courses and trying to predict whether this student is going to fail or not. The demographics are used mostly just [...] race or ethnicity, which doesn't change across time. But there're a lot of factors that actually change across time that needs to be taken into account. For instance, financial aid. They might have financial aid in their first semester, they may lose it if they don't have [a high] enough GPA.''
% \end{quote}
For example, when machine learning systems are trained on academic and curricular data alone, socioeconomic factors such as student finances and family are left out of the bigger picture of student success. Commenting on an ML4Ed paper on college student performance prediction, P02 (Director of Research) says she ``feel[s] a little uncomfortable thinking that graduation time and graduation grade is entirely dependent on course work which is really not true for college.'' Instead, there are institutional decisions such as how much aid students receive each semester and macroeconomic conditions that could induce changes to familial financial circumstances. The models do not account for factors that relate to institutional choices that change over time and impact the likelihood of student success; yet non-malleable factors related to student race are taken into account.
In the following section, we examine the use of demographic information as input features, within the broader context of education equity.

%A significant portion of the bigger picture of socioeconomic data (e.g. finances, family) is not collected. This weakens systems trained on curricular data alone. Choices of what to take into account - broader than individual sphere. Problem formulation favors individualistic vs structural/social view.

\subsection{Designing inputs with education equity in mind}

Educational access and equity are primary social goods. While several participants emphasized the importance of acknowledging and addressing structural inequality in education-related research, the ML4Ed papers rarely explained how their research understands existing education inequities. ML technologies that are agnostic to structural inequality, such as the systematic disadvantages faced by under-resourced institutions, can actually widen gaps in access to quality education (see Section~%\ref{sec:misuse_grading}
1.B in the supplementary text).
% \begin{quote}
%     P12 (Assistant professor): ``I'm always attentive to how research agendas are shaped that take into account the incredible structural inequalities in the United States and around the world. And is there a way that interventions are -- is there way that people's research agendas are paying attention to those structural inequalities?'' %And this sort of seems to be attempting to do so but in a [...] naive kind of way.
% \end{quote}

\paragraph{Unconsidered use of race data for prediction}
Structural inequality manifests as disparities in educational opportunity and advantage across demographic groups. When it comes to ML4Ed research, however, the use of demographic information should be approached with care, according to several participants. 
  \begin{quote}
      P04 (Senior lecturer): ``[We] would be sending the wrong message if we said, we were controlling for race and ethnicity, because it might imply that we thought that kids who were Black or Hispanic [...] were less likely to succeed.%[...] correlationally on average that is true, but it is not true of every kid.
      ''
  \end{quote} 
  Commenting on a paper that included race as an input feature for a machine learning model that predicts student academic performance, P04 warns that this choice has unexamined normative implications, such as sending the ``wrong message about how we assess risk''. 

Even if the inclusion of  attributes led to increased predictive accuracy (when measured on a particular dataset), such nominal improvements must be weighed against how the data is used and interpreted and the broader implications of the decision to explicitly include, and therefore, reify, race as an input to the predictor. Importantly, including race and ethnicity, without accounting for the ways in which resources and opportunity are allocated by race and ethnicity can lead to erroneous conclusions about the role of race and ethnicity in outcomes. 
This decision sets a standard for how race is understood and instrumentalized in that education context---which can be problematic. In the following excerpt, P04 contrasts a state agency's ultimate decision to exclude race from \emph{their} predictive model with the ML4Ed paper's approach.

\begin{quote}
     P04 (Senior lecturer): ``[Race] was a little too nuanced and it wasn't adding enough predictive validity to make it worth the cost of the potential for people being up in arms about the state's way of doing these models. [...] But a researcher would never think of it that way, right? They [...] want to get the best prediction possible, and you probably would add a little bit to your predictive ability if you had that in there.'' 
     % But they're not the ones who have to explain to a bunch of irate guidance counselors.''
\end{quote} 
\noindent If the researcher only considers the predictive power of the race feature without thinking about the social implications of including it, this may lead to negative consequences when the model is used.

Even if race were included as an input to a model, it can potentially have a large measurement error. To illustrate one of the many complications, P14 (Postdoctoral researcher) points out that there can be ``different race and ethnicity variables across the different datasets''; not all demographic information that agencies `collect' is self-reported. Self-reported racial and ethnic data could significantly diverge from data collected by government agencies.

%Although demographic information is often contentious as a model input, participants acknowledge the utility of having demographic data and other information about the student---in the context of student success---for identifying systemic problems and loopholes, when used in combination with local expertise and on-the-ground knowledge (P04).

\paragraph{Auditing for gaps in educational access}

In the context of university admissions, demographic information may be used to audit the existing or proposed admissions process for gaps in education access. ML4Ed papers often report feature weights and other feature importance measures for demographic input features, but these metrics are in general insufficient for assessing whether a proposed algorithmic system is supporting or undercutting ``big picture'' equity goals. For example, one ML4Ed paper on graduate admissions found that their predictor of admission probability placed zero weight on the demographic attributes of applicants, including gender and ethnicity, and concluded that ``admissions decisions are based on academic merit''. P06 (PhD candidate), however, points out that this link is tenuous: ``it's better than the worst case although we know [gender, ethnicity and national origin] were encoded in other non-demographic features as well.''
% \begin{quote}
%     P07: ``When you look at your equity data big picture, how many students do you have enrolled across different disaggregated demographic factors?''
% \end{quote}
Instead of uncritically interpreting feature weights, participants suggest looking directly at statistics such as student enrollment rates for a realistic picture of education access.

 By giving consideration to education equity from the beginning---when formulating education goals into ML problems---researchers can work to address the real and pressing problems of equity and access in education. 
 In all, problem formulation choices in ML4Ed are often under-determined, yet many have far-reaching consequences. 
 
\section{Translating Predictions to Interventions}\label{sec:interventions}

We've shown how effective translation from education goals to problem formulation choices can be challenging, especially in settings with multiple stakeholders and long term goals. However, the ML life cycle doesn't end here---we next discuss how thinking critically about interventions and impact can further guide key design decisions and research directions.

\subsection{The gap between predictions and interventions}
One of the most common and consequential gaps that participants identified in the ML4Ed papers is the gap between prediction tasks and real-world interventions.
%Participants concurred that almost no problems in education are ``pure'' prediction tasks, and that ML researchers interested in improving outcomes should pay attention to where their model fits in an intervention pipeline, and to the eventual validation of the impact of their models.
\paragraph{Predictive accuracy is no panacea} Instead of focusing on interventions, most machine learning papers tout improvements in prediction accuracy---that is, how often the predictions turn out to be `correct,' by some measure---whether it's predicting student dropout risk, predicting test question difficulty, or predicting likelihood of admission. However, improvements in prediction accuracy do not automatically translate into improved outcomes, and in fact may not affect outcomes at all. This gap surfaced in the majority of interviews (12 out of~15).

\begin{quote}
    P12 (Assistant professor):  ``[Researchers] are trying to make tiny, substantively completely irrelevant improvements in the area under the curve of some prediction algorithm, `can we predict 62\% instead of 61\%,' but nobody had anything to do that would actually help these people.''
\end{quote}

\begin{quote}
    P13 (Full professor): ``You don't improve things by predicting them better. There is a missing link there obviously between how we act on predictions in social spaces that are incredibly complicated to improve outcomes.''
\end{quote}

\noindent In the risk prediction setting, P09 (Assistant professor) and P08 (PhD candidate) warn that even if a prediction of student dropout risk is highly accurate, without adequate resources to help those students, this information ultimately does not benefit institutions.

\begin{quote}
    P09 (Assistant professor): ``Even if you tell [schools] 
    %in percentage based on the student's attributes and based on the student's background, based on the student's achievement 
    [...] that [a student has] a 97\% chance of dropping out based on our training data, that's a difficult thing to take in especially in the public schools [where it is] very difficult to find good teachers for those students.''
\end{quote}

\begin{quote}
    P08 (PhD candidate):
    %Schools do a lot,
    ``A lot is put on schools, a lot is put on teachers. [...] An unfunded mandate on teachers and what they are expected to do in the classroom [with a risk prediction tool] could be bad.''
\end{quote}

\noindent Thus, the `accuracy' of risk prediction is not a driving force for improving student outcomes, as it does not provide a means to improve student outcomes.

\paragraph{Validation within an intervention pipeline} If predictions do not directly translate into interventions, what is the value of improving them? Supposing that there is eventually an intervention in mind, and that the prediction is a concrete component of the intervention pipeline, then P13 (Full professor) comments that improvements in prediction quality can lead to incremental improvements in the efficacy of the known, larger intervention pipeline. Mistakes to avoid, however, include ``overselling'' the contributions of the ML component, or forgetting that additional data driven methods will be required to validate the full intervention pipeline.

Six participants highlight the importance of validation in the translation from prediction to interventions. 
%One important reason that ML researchers should be aware of where their proposed predictive model fits into an intervention pipeline is to discuss the the validation of interventions that use their proposed predictive models. 
P13 (Full professor) emphasizes that education practitioners will ``ultimately still have to test down the line'' the impact of the intervention. Similarly, P02 (Director of Research) comments that ``the next step after prediction'' is a causal inference problem testing the efficacy of the downstream intervention. To even begin to measure the effect size of an intervention, ``you need to have an intervention in mind'' (P02, Director of Research). While causal A/B testing is common practice in industrial applications of ML, many ML4Ed papers do not reference this need. 

\paragraph{Forward-looking policy recommendations} As ML researchers explore new techniques, intervention pipelines that utilize these new technologies may not yet be well established. If the intervention pipeline isn't established, can such research and development of predictive models still add value? Perhaps this is possible, especially if ML papers include downstream consideration of interventions and policy recommendations, a practice that is standard in, for example, quantitative subfields of economics. After all, when ML papers already reference policy-driven motivations such as improving student financial situations, it would be remiss not to include a discussion of downstream interventions stemming from this motivation, as P02 suggests below.

\begin{quote}
    P02 (Director of Research): ``It's not about just predicting and identifying who is going to be at risk...we should think about, `is the financial aid program that we're offering the right thing to do or do we need to make changes there?'''
\end{quote}

% As another example, predictions of student dropout risk can add value through the identification of individuals in need of intervention.  

% \begin{quote}
%     P02 (GSE Faculty): "Prediction really helps us identify who is at risk, that's important. Now, with the people who we identify as at risk, now we have to think about what are we going to give them."
% \end{quote}
\noindent Devising new intervention pipelines and policy recommendations that effectively bridge the gap between prediction and intervention is likely to require further interdisciplinary collaboration between education practitioners and ML researchers. 

%This important exercise would benefit both parties, with ML researchers achieving greater impact and adoption of methods, and the education field receiving more conscientiously developed technologies.

\subsection{Harms of naive translation from prediction to intervention}\label{sec:interventions:harms}
%Harms of naive translation from prediction to intervention

When ML researchers do not engage with education experts to thoroughly consider interventions, this can lead to naive application of predictive models that cause unintentional harm. In this section, we focus on the topic of risk prediction, and leave discussions related to automated grading and tutoring to the supplementary material.

\paragraph{Potential harms of student risk scores} According to five participants, the translation of predictions to interventions is a particularly salient issue in early warning systems applications for high schools, higher education, and MOOCs, where ML4Ed papers proposed new models for predicting individual student risk, e.g., of dropping out of a course or not graduating on time. A `failure' event easily translates into a concrete binary outcome, making it a common prediction target for ML papers.

Unfortunately, the ease of quantifying the target does not directly translate into effective interventions. P12 (Assistant professor) points out that one naive assumption that ML papers tend to make, either implicitly or explicitly, is that ``showing people more data is good.'' Some ML papers claim that they may ``[present] at risk students with meaningful probabilities of failure.'' However, the efficacy of sharing risk probabilities with individuals is not grounded in behavioral research. 

\begin{quote}
    P12 (Assistant professor):  ``The odds that [presenting at at risk students with probabilities of failure] is going to be helpful for students just seem so phenomenally low to me. Have the authors of this paper...seen any evidence talking with students who are in a kind of marginal state? [...] It doesn't seem well connected to research about what motivates people.''
\end{quote}
In fact, directly showing risk scores to students, teachers, or administrators can worsen student outcomes. If a system categorizes students as ``at risk,'' then P01 (Executive Director and Professor) points out that ``there is a tendency for a lot of these systems to stigmatize students.'' Such stigma from a ``deficit-minded'' classification can demotivate students (P10, Senior director).

% \begin{quote}
%     P10 (Director of Student Success Analytics): If you sit somebody down and say, ``You're at risk,'' and use this kind of very deficit minded language, [...] it may not motivate some students, but it's probably going to demotivate other students.
% \end{quote}
Even if a system is only implicitly categorizing a student as at risk, students are sensitive to differential treatment and cues alluding to their abilities.

\begin{quote}
    P06 (PhD candidate): ``If a tutoring algorithm systematically underestimates female students' mastery levels and provides them with instructional sequences or feedback messages for struggling students, some female students might question their own abilities which could decrease their motivation. Eventually this might lead to a self-fulfilling prophecy.''
\end{quote}

%Other adverse reactions are possible. 
% Instead of internalizing the risk prediction, students and families may also view the failure risk as being of no value. At best, this leads to the student ignoring the risk categorization; at worst, 
% Given the stigma of these deficit-minded classifications, the use of predicted risk indicators can strain the relationships between schools and families. %and washing out any benefits of the predictor. 

% \begin{quote}
%     P14 (Postdoctoral researcher): ``When you have predicted indicators of not doing well or falling behind in education, it's always a tricky thing to bring to students or their families, because [...] [these] can often make families feel defensive, or feel like you are not looking at their child as an individual, or you're not understanding the particular situation that that family is in.''
% \end{quote}

\paragraph{Institutional level harm} In addition to influencing students' views of themselves, naive application of student risk scores can also negatively influence the the students' support structures, including teachers, parents, and institutions. 
% For teachers, the same risks apply of either over-indexing on the risk score and harmfully misjudging a student, or simply dismissing the score entirely if it doesn't align with their firsthand observations of the student.
If shown risk scores directly without any additional guidance, a teacher ``might allocate more of their limited time to other students rather than a student that the model seems to predict that they will not graduate'' (P09, Assistant professor). Similarly, ``a parent might stop investing in that child or spending as much time with them'' (P09). 
If risks scores are shown at the institutional level, this can lead to allocation of resources that align more with institutional incentives than student well being.

\begin{quote}
    P08 (PhD candidate): ``If schools [focus] too much on what [they] think students can achieve, [they] end up putting in these artificial barriers and filtering [students] in ways that [they] think are going to work for them or more pessimistically work for the school.''
\end{quote}

At the institutional level, P08 (PhD candidate) further points out the potential for additional ``surveillance of the students who are at risk,'' possibly without the student's knowledge or consent. Thus, students may not be comfortable with the knowledge that their data is used for risk prediction. P01 (Executive Director and Professor) expresses similar reservations even if the risk scores are withheld.

\begin{quote}
     P01 (Executive Director and Professor): ``If we had an ordinal ranking of students by risk and so forth, I would not be comfortable with sharing that with students nor would I be comfortable with saying, `we have it but we're not going to tell you.'''
\end{quote}

% With institutional incentives in mind, P14 (Postdoctoral researcher) suggests that ML practitioners formulating the input data and risk prediction task should factor in any biases in how the model will present the students to agencies that are applying interventions using these risk scores. 

% \begin{quote}
% P14 (Postdoctoral researcher): ``If your goal is to create an intervention for this agency, then you want to understand from their viewpoint how will the student look, regardless of whether or not that's the best quality data across all options.''
% \end{quote}

% Even with these considerations of bias  and discernment in when and to whom risk scores are shown, Fundamentally, P08 (PhD candidate) points out that students still may not be comfortable with the knowledge that their data is used for risk prediction, ``especially if it's more visible to the world.'' P08 (PhD candidate) further points out the potential for discomfort with additional ``surveillance of the students who are at risk.'' P01 (Executive Director and Professor) expresses similar reservations even if the risk scores are withheld.

% \begin{quote}
%      P01 (Executive Director and Professor): ``If we had an ordinal ranking of students by risk and so forth, I would not be comfortable with sharing that with students nor would I be comfortable with saying, `we have it but we're not going to tell you.'''
% \end{quote}

\noindent The publication of ML models for risk prediction can endorse naive interventions with risk scores if those same papers do not discuss the downstream usage and potential harms. The responsible and beneficent use of risk scores in the education sphere is still an open problem for future ML4Ed research.

\subsection{Towards more intervention-aware predictions}% While the naive application of predictive models can lead to harm, 
We now outline ways that prediction tasks can be better formulated with interventions in mind. This includes participants' existing success stories and suggestions for future work.

\subsubsection{Need for Actionability (vs. Interpretability)}\label{sec:interventions:actionability}

To mitigate harms from the naive announcement of predicted risk scores, participants repeatedly pointed to the value of building models that provide actionable insights, where ``actionable'' generally refers to the ability to take helpful actions. In the risk prediction setting, P10 (Senior director) recommends stepping away from the ``deficit language'' and ``focusing on the ways in which the student can move forward positively, and hopefully get to graduation.''

\begin{quote}
    P10 (Senior director): ``We don't have to say, `You're not going to succeed.' We can say, `Let's talk about what are the decisions that you need to make, what is the pathway forward that will allow you to succeed.'''
\end{quote}

\noindent P03 (Associate professor) notes that the risk score is a blunt instrument that students can't directly use to improve outcomes. 

\begin{quote}
    P03 (Associate professor): `` `You're in the 10th percentile for something' sounds different than `we're worried because you've been absent a lot.' ''
\end{quote}

\noindent Actionable insights on the path toward success, such as telling a student that the absences may hurt their class performance, can lead to clearer interventions. 

While many ML papers included analysis of feature importance in their risk prediction models, this version of \emph{interpretability}---making a machine learning model more understandable to a human~\citep{doshi2017towards,lipton2018mythos}---fell short of addressing experts' needs in providing \emph{actionability}. %\lnote{To distinguish from interpretability, we can this ability to ``most interested in actually what we could do'', i.e. define our use of actionability in a short phrase. } 
P04 (Senior lecturer) notes that feature importance analysis of non-mutable traits such as demographics is not always useful for developing interventions, and may distract from the analysis of more actionable behavioral factors.

\begin{quote}
    P04 (Senior lecturer): ``What is most interesting about it to me is not, `I wonder if the demographic factors matter more than the behavioral factors.' To me it's more about, `what can we actually do to help kids get off the trajectory they're on if they're not on a good trajectory.' ''
\end{quote} 

\noindent P04 further joins two other participants in pointing to causal evidence as a useful tool for showing the value of behavior changes. We discuss the connection to causal inference further in Section~\ref{sec:action-causal}. %Emerging work in causal inference suggests ways to develop intervention models as opposed to prediction models in various domains \citep[e.g.][]{motz2018embedding,prosperi2020causal}, but existing causal inference approaches have limitations of their own \citep{Stephenson98RCT,deaton2018understanding} and identifying actionable insights in a data-driven context is still an open challenge.

% Still, harnessing ML to provide actionable insights is an open problem which has seen relatively less research investment compared to optimizing predictive power. 

% quantifying the relative feature importance and exact causal effect sizes is not always necessary.

% \begin{quote}
%     P04 (GSE Faculty): ``We want the kids to show up for school and it doesn't matter if we get the exact right answer on how much it will increase their likelihood...Are they going to do anything differently if you tell them that they have to get to school three more days a month or 10 more days a month? Probably not.''
% \end{quote}
% Although measuring causal effect can be difficult from a methodological standpoint, P04 notes that the problem is actually often simpler: knowing the direction of the causal effect already adds value when it comes to improving outcomes.

\subsubsection{Design for empowered human operators} In tandem with actionable insights, five participants highlighted the importance of designing ML to \textit{empower} human operators such as academic advisors, teachers, administrators, and admissions officers. Going beyond simple corrective actions, these operators may use ML tools to amplify their ability to achieve broader education goals.  
%Their involvement goes beyond simple corrective actions. Instead, the human operator is \emph{empowered} to use the ML model as a tool to amplify their ability to achieve broader education goals. 

\paragraph{Centering the role of teachers} In a classroom setting where the teacher is a central human operator, predictive models can improve the teacher's capacity to provide individual student attention and reduce their workload. For instance, automated tutoring tools can effectively reduce the burden on the teacher as a distributing authority for information. 

\begin{quote}
    P11 (Associate professor): ``First you look online or you ask a friend; if that doesn't help, then you ask another friend; if that doesn't help, then you go to the teacher and so that way the teacher is sort of able to distribute themselves a little more evenly.''
\end{quote}

% At their best, alerts from risk prediction tools can also help teachers as long as the teachers have ``enough power to make any changes and know what to do'' with the alert (P02, GSE Faculty).
\noindent When the teachers have the power and resources to intervene, risk prediction algorithms can also ``help [teachers] catch some kids that maybe had fallen below the radar before and giving them another source of data on that'' (P04, Senior lecturer). Key to this statement is that the algorithm is just acting as ``another source of data'' in tandem with the teacher. The goal of identifying students that the teachers would otherwise have overlooked is different from the standard goal of achieving high accuracy for the whole population of students. This suggests that the value of risk prediction systems would be improved if they were designed with the teacher's partnership in mind.

\paragraph{Centering the role of academic advisors} In addition to teachers, participants highlighted academic advisors as key human operators that ML systems can aid. When it comes to student advising, instead of thinking, ``this technology will solve our problem'', P10 (Senior director) prefers the attitude, ``this technology will be a tool in our toolkit while we do our job that will [...]
%then help to 
hopefully have a positive impact on students.''

% \begin{quote}
%     P10 (Director of Student Success Analytics): ``[O]ur advisors work harder now than they did 10 years ago, before we had any of this technology, because they're balancing the needs of their caseload of students and the conversations that they're having, with the alerts that they're getting from the system and the conversations that they have to have and all the tracking.''
% \end{quote}
% In P10's experience deploying data driven tools for advising, the deployment of the tools doesn't actually lead to any less reliance on advisors in the advising process. Instead,  the advisors are working harder than in the past and the technologies help them manage their extensive student caseloads.

A key situation when an ML system falls short alone but works well in partnership with a human advisor is when there is ambiguity in the student's needs, and students ``don't know what they don't know.'' Effective usage of automated chatbots or search engines requires knowing what to query; however, P03 (Associate professor) states that ``where students get held up is not knowing the questions that they need to ask.''

\begin{quote}
     P03 (Associate professor) ``If I don't know the specific terminology that's used at my school, or if I don't know how to think about like a certain question, or if I can articulate my goal but I don't have a knowledge of all the different paths that could get me to that goal, I think that's where [...] technology driven advising solutions can't advise students as well.''
\end{quote}

Instead, in between the specific questions that students can ask chatbots or search engines, advisors are ``able to fill in the gaps for students'' to help them ``envision a pathway'' from a relatively vague conversation about their broader goals. P10 (Senior director) provides a specific example of an advisor guiding a student to such a point when the student only provides vague guidance on their goals: 

\begin{quote}
    P10 (Senior director): ``Instead of saying, `You're not going to be a nurse, sorry, like, good luck,' it's more, `Well,  [...] we have respiratory therapy, or we have nutrition, or we have bioinformatics. We have all these other healthcare disciplines that might allow you to help people, to work in healthcare, and to get a job, which [are] three boxes you said you wanted to check. And in some cases, the students are like, `That sounds great. I had no idea what an occupational therapist even was.'~''
\end{quote}

According to P07 (Vice president), one obstacle to realizing these benefits of pairing technological tools with dedicated advisors and teachers is that ``often machine learning is really hidden within an ed-tech tool.'' This means that users and administrators alike ``don't understand all the places where [ML] actually is embedded now.'' For example, in admissions, a predictive model may be embedded so deeply in the decision pipeline that admissions officers or the higher level administrators are not fully cognizant of its role. The lack of transparency coupled with limited user technical knowledge makes it difficult for a human operator to audit or modify the usage of ML within these systems, including overriding incorrect predictions or providing feedback to improve the ML models. This underscores the need for ML systems and pipelines that are designed with the empowerment of human operators in mind. %The issue of transparency, at least, falls on ML system designers.

\section{Discussion and Related Work}\label{sec:dis}

Machine learning promises automated procedures that recognize meaningful patterns in education data, and provide principled, real-time decision support and interventions to improve educational outcomes \cite{kotsiantis2012use,Liu2017datadriven}. However, before any of these promises can be realized, practitioners and researchers must traverse the entire machine learning life cycle (Figure \ref{fig:process}), from goal identification and problem formulation to intervention and impact evaluation. Through our qualitative work bringing the expertise of education researchers to bear on the research practices of ML4Ed, we found that varying levels of attention have been paid to different parts of the life cycle.

In Section \ref{sec:problem_formulation}, we showed that the consequential choices in the translation of education goals to machine learning tasks are currently overlooked. We found that multi-faceted education goals are often reduced to a single quantitative metric, while expedient choices in prediction target and input lead to the omission of key education goals, such as education access and equity, and stakeholder interests. In Section \ref{sec:interventions}, we outline gaps in the translation from prediction tasks to interventions, including negative externalities of naive application of predictive models, and a path forward from education researchers towards formulating more intervention-aware prediction tasks. Taken together, this work contributes insights to the ongoing conversation around machine learning and its impact in education, as well as broad, cross-domain critical discussions at the intersection of algorithmic fairness, accountability, transparency and ethics (FATE) and AI for Social Good (AI4SG). Situating our findings in FATE, and related fields, we discuss both shared insights and new tensions that emerge from our work. %\todo{summarize two aspects where our findings break with or add a unique perspective to existing literature}

%[outline major findings \todo{}] The previous two sections have provided a detailed expansion of the view of the ML life cycle to include problem formulation (Section \ref{sec:problem_formulation}) and the translation from predictions to interventions (Section \ref{sec:interventions}). These sections highlight aspects of these phases that are currently overlooked, consequences and harms that come from overlooking them, and opportunities for ML researchers to contribute in these phases that can lead to better outcomes.

\subsection{Related work in education technology}\label{dis:edtech}

The educational data mining (EDM) \cite{labarthe2018analyzing, calders2012introduction}, learning analytics (LA), Computer-Supported Collaborative Learning (CSCL) \cite{lin2018examining}, and AI for education (AIED) \cite{chen2022two, pinkwart2016another} communities have been active for the last one to three decades. Retrospective surveys of the research trends and ongoing challenges of these fields have analyzed common and growing ML paradigms being applied in education, like neural networks for learning characteristic prediction and teacher evaluation \cite{chen2022two,tang2021trends}, NLP for language education \cite{chen2021twenty}, and AI-assisted personalization \cite{chassignol2018artificial, zawacki2019systematic,chen2022two}. Our contribution to this discourse is in discussing \emph{how} and \emph{why} these different ML paradigms are chosen, with a target audience of ML researchers who are invested in developing improvements on top of these paradigms or novel paradigms entirely.

% \swedit{This work focuses on contributions to the education domain from broad machine learning and AI conferences such as KDD, AAAI, and IJCAI, since 2013. These have a more general technical audience than the many education technology conferences, for which ``there have been tendencies towards smaller, more focused conferences and communities'' \cite{pinkwart2016another}. Several works provide comparisons between these communities, including philosophical differences and differences in interdisciplinary engagement \cite{pinkwart2016another, rienties2020defining, guan2020artificial}. 
% % \cite{guan2020artificial} provide a more detailed survey of research themes and historical trends in papers applying AI to education published in journals.
% Our interview study adds to this discussion by providing a deep critical dive into works published in ML/AI conferences, which complements other broad surveys and historical comparative analyses such as that of \cite{guan2020artificial}.}

Based on the findings of this study, we reaffirm and provide further technical commentary on some of the challenges presented by education technology research communities. For instance, the AIED retrospective by Chen et al. \cite{chen2022two} highlights the challenge of adoption by human operators, and propose ``exploring how human and automated instruction can most effectively be combined to best support instruction'' as a future research direction. We show this to be a challenge with publications in general ML conferences as well, and further propose several ways that the ML research community can approach this issue. In particular, we suggest focusing on \emph{actionability} and formulating ML problems to \emph{empower} human operators.

The distinction between prediction and intervention, as well as the challenge of designing interventions with predictive models, has been acknowledged in prior work in AIED and LA, particularly in context of dropout prediction \citep{campbell2007academic,xing2019dropout,tsai2020precision} and predictive learning analytics \citep{baker2011towards, hlosta2021impact,karumbaiah2018predicting,wong18learning}. Still, many works from the EDM and AIED communities focus on predictive performance \cite{lin2017comparisons, karumbaiah2018predicting, beaulac2019predicting,aulck2019mining, chen2022two}, with less discussion of whether or how the improvement in predictive accuracy translates to better outcomes. A number of recent empirical studies \citep{dawson2017from, borrella2019predict} have found the effect sizes of targeted interventions based on student risk predictions to be statistically insignificant.
Our work builds upon these observations to further problematize the assumed relationship between `accurate' predictions and beneficent interventions, which, as far as we know, also exists in the education technology community \citep{chatti2013reference}.

%\llcomment{add 4 references about ed work drawing distinction between prediction and intervention in drop out prediction. also acknowledge the challenges. AIED papers might be more aware of the gap, reference/surface the problem, but we illuminate the particular ways in which prediction and intervention gap is not acknowledged in ml4ed, and we discuss the problem at more depth. new points about validation and does improving accuracy help? also applies to the AIED work. examples of drop out prediction and }

Recent work in the intersection between education technology and FATE has addressed the impacts of education technology on equity and ethics \cite{madaio2021beyond, holstein2021equity, holmes2021ethics}.  Madaio et al. \cite{madaio2021beyond} apply a critical theory lens to evaluate the impacts of education technology and algorithmic fairness notions on education equity. Holstein et al. \cite{holstein2021equity} analyze ways that ``AIED systems'' may alleviate or amplify inequalities under current practical usages. Holmes et al. \cite{holmes2021ethics} set a framework for the ``Ethics of AIED'' through a survey of researchers that publish in the journal and conference of Artificial Intelligence in Education. Our work adds to this discourse in three ways. First, we focus more broadly on impact than on ethics and equity per se.
Second, instead of interdiscplinary AIED communities, we target technical ML communities that are potentially driven more by algorithmic novelty than societal needs \citep[see related,][]{birhane2022values}. Finally, our methodology of studying the disciplinary boundary crossing of ML communities through interviews with education experts provides an additional distinct evidentiary lens.

Given our different audience and broader focus on impact, our study consequently foregrounds a different set of questions and recommendations than the ethical framework of Holmes et al. \cite{holmes2021ethics}. For example, Holmes et al. \cite{holmes2021ethics} discuss 
the  ``value of transparency'' as motivated by policy, but does not mention \emph{actionability}, which is a more impact-driven desideratum. Holmes et al. \cite{holmes2021ethics} also extensively discuss issues of data governance, such as privacy, anonymity, ownership, and control. These critical ethical problems are much better addressed from their interdiscplinary policy lens, and transcend the technical ML problem formulation choices that we discuss.
% Differences in topic focus:
% Holmes discusses the "value of transparency," but does not talk about actionability.  

% Data governance: privacy, anonymity, ownership, control. These are policy problems that can transcend technical ML modeling choices. 

% Holmes talks about the possibility of replacing human teachers and the ethical issues surrounding this. We also discuss the empowerment of human operators, but do so from a lens of impact rather than the broader ethics of replacement. 

Despite these differences in focus, we also view equity as integral to evaluating choices in problem formulation and interventions, and many of the recommendations from these works align with ours. Madaio et al. \cite{madaio2021beyond} note that education AI technologies ``are forged in historical relations of power,'' and ``may reproduce structural injustices — regardless of the models' accuracy or fairness.'' Holstein and Doroudi \cite{holstein2021equity} discuss disparities in access and usage of education AI technologies. Both of these effects can lead to the types of gaps between prediction and intervention surfaced in our interviews in Section \ref{sec:interventions}. 
% Holmes et al. \cite{holmes2021ethics} also notes that the AIED community ``needs to learn to integrate fairness-promoting algorithms into the system.'' We take this a step further by arguing that the ML communities developing "fairness-promoting algorithms" need to do so with downstream impacts in mind, which requires accounting for the planned intervention in the definition of a ``fairness-promoting'' algorithmic objective.
Our work further joins Madaio et al. \cite{madaio2021beyond} in concluding that ``quick technical solutions and neat group-level evaluations of `AI
fairness'~'' are not enough to produce adequate solutions to complex issues of education equity. The proposed extended ML life cycle (Figure \ref{fig:process}) constitutes our approach for illustrating the limitations of focusing on ``neat'' technical solutions to equity challenges. Finally, a common recommendation shared by all of these works is the importance of focusing on the human operator: Holstein and Doroudi \cite{holstein2021equity} highlight the importance of AIED systems' ability to communicate limitations and ``hand off control to humans,'' and Holmes et al. \cite{holmes2021ethics} discuss ethical issues of human agency. Our work adds to this discussion with an angle of providing specific recommendations for how ML researchers can empower human operators through problem formulation choices and development of better-targeted %underappreciated
methodologies.
% \cite{madaio2021beyond}'s discussion of \textit{design justice} also relates closely to our discussion of problem formulation, especially if equity and justice are not included in the translation between education goals and machine learning objectives.

% Holmes points:

% Ethics in AIED attempts to create "an actionable code of best practice that the community can rely on in designing and deploying AIED technologies in diverse educational contexts." Our goal is similar, but aimed at a different community. While Holmes's code of best practice would be primarily aimed at those choosing what to implement in different learning settings, we want to provide a code of best practice for those inventing new techniques that may be implemented someday in the future. We look at gaps in understanding in the ML development process from the AI/ML research communities, which generally focus on inventing novel algorithmic methodologies, and are driven by algorithmic novelty. This primary driver of algorithmic novelty is less present in AIED, which focuses more on novelty in teaching/learning methologies and principles. 

% Holmes: AIED IS "primarily led by technology", noting that "its impacts on stakeholders, including direct users, those indirectly affected and society at large has been largely ignored." 

Beyond these discussions of education equity and ethics, our work also connects with other styles of argument in FATE more broadly, which we discuss in the next section.

\subsection{Related work in FATE}

Emergent critical scholarship in FATE have pointed out the \emph{gap} between the abstract goals of computational research and system design \emph{and} their operationalization, often in a sociotechnical system   \cite{passi19problem, blodgett2020language,jacobs2021measurement}. Past studies have diverse methodological approaches, scopes, and abstractions to guide their critical inquiry \cite{dean2021axes}. Measurement modeling from quantitative social science suggests that many of the harms of computational systems discussed in the FATE literature \cite{Barocas16} can be traced to mismatches between ``unobservable theoretical constructs'' and how they are ultimately reified as measurements \cite{jacobs2021measurement}. Our work finds there to be varying degrees of mismatch between abstract education goals and their operationalization in the extended machine learning life cycle, going beyond the issue of measurement to, for example, question the theoretical constructs themselves. By highlighting the selective interpretation of education goals, our findings suggest that the choice to focus on a single theoretical construct, particularly narrowly defined constructs, such as ``strength of candidate'' in the context of graduate admissions, already sidelines certain education goals, often, educational access and equity.  %For example, in the goal identification and problem formulation stage,  as a single quantitative metric, but also their operationalization as prediction-driven interventions.  

Our work echoes a key insight of a recent ethnographic study on the formulation of corporate data science problems that ``problem formulation is a negotiated translation'' and has normative implications \cite{passi19problem}. The results in Section~\ref{sec:problem_formulation} highlight the normative implications of problem formulation in the education context, where the goals of employing data science and machine learning are arguably more nuanced and multi-faceted than in the corporate context, and the social stakes higher. By contributing a distinctly interdisciplinary perspective grounded in expert knowledge of the education domain, our work builds on both the critical scholarship and the body of practical guidance for negotiating \emph{translational gaps} when machine learning is applied to consequential domains.

In terms of practical guidance, a pioneer study on racial bias in health risk prediction surfaced the issue of label bias and advocated for more careful choice of the prediction target as a way to mitigate racial disparities in predictions (i.e., recommending additional care for Black patients at a lower rate than for similarly ill White patients) \cite{Obermeyer19dissecting}. In our study, participants also pointed to the risk of choosing prediction targets that are actually proxies for immutable socioeconomic and demographic factors. Our findings support paying greater attention to the choice of the prediction target in ML4Ed problem formulation, but the reasons go beyond the mitigation of prediction bias. By highlighting other complications such as the loss of non-quantitative information, the alignment with education goals and needs of stakeholders, and threshold effects, our work provides a broader account of the important factors that should go into the choice of a prediction target.

%paragraph 2:\todo{Lydia} connection to the label bias work.  Beyond bias: a lot of work in this space has focused on the idea of bias, data bias and label bias. discussion of how our work goes beyond thinking about bias (which doesn't necessarily question the choice of a single quantitative metric, interventions, etc.) to thinking about problem formulation and interventions/impact \cite{Obermeyer19dissecting}

% Conclusions from our interviews provide suggestions to broaden the scope of considerations in the ML design process. For example, beyond just focusing on improving predictions, ML researchers would have more impact if they incorporated the translation of those predictions to interventions in the problem formulation process.

\subsection{Connections with HCI}

This broadening of scope and critical reflection of the design process also has a long history in the Human Computer Interaction (HCI) and Science, Technology and Society (STS) literature, where concepts of \textit{reflective design} \cite{sengers2005reflective}, \textit{participatory design} \cite{schuler1993participatory}, and \textit{value sensitive design} \cite{friedman2013value, zhu2018value} provide frameworks to bridge the gap between designers, users, and their implicitly held values in the design process. Shared by these design frameworks is the idea that a computational system designer's choices embed implicit assumptions and values, and in order to best serve the user, the system designer must incorporate the evaluation of these values into the development process. Our work brings similar critical reflection to research papers applying ML in education. Just as the system designer's scope extends beyond implementation to also incorporate the values and needs of the user and societal context, so we also find that ML researchers could have more positive impact in the education sphere if they incorporate critical evaluation of values and interventions into the choices made in problem formulation. 

% Regarding the translation of predictions to interventions, 
The HCI literature on human-centered algorithm design \cite{baumer2017toward, saxena2020human} relates closely to our discussion on the translation from predictions to interventions through human operators, and includes case studies from outside of the education domain such as the U.S. criminal justice system  \cite{green2019disparate} and the U.S. child welfare system \cite{saxena2020human}. Human-centered algorithm design goes beyond requiring human oversight of algorithms, which can be limited or ineffective \cite{green2022flaws}. Floridi et al. \cite{floridi2020design} posit the importance of 
respecting user autonomy and ``optionality'' through ``receiver-contextualised intervention'' in AI for social good projects. %In the case of student risk prediction, this could mean asking if teachers retain the genuine option to ignore recommended algorithmic scores. 
We join these works by providing examples and insights grounded in the education domain (e.g. the partnership with teacher-advisors) that speak to the value of \emph{empowering} human operators more systematically.

\subsection{Actionability and causality}\label{sec:action-causal}

Our findings around interventions also contrast the clean division between causality and prediction problems drawn by Kleinberg et al. \cite{kleinberg2015prediction}. 
%Through our interview study and ML life cycle framework,
% We find the lines to be much more blurred in the education domain, as the subsequent intervention that uses a given prediction is key to the efficacy of the predictive model. 
The framework in \cite{kleinberg2015prediction} does not explicitly recognize the translation from prediction to intervention; however,
%and is thus limited in its ability to categorize ML problems applied to education. In fact,
our findings may be used to extend the framework by Kleinberg et al. \cite{kleinberg2015prediction} to more thoroughly evaluate the practical efficacy of prediction tasks. To illustrate this, consider the following example of a ``pure prediction task'' from \cite{kleinberg2015prediction}: ``\textit{The policy challenge is: can we predict which surgeries will be futile using only data available at the time of the surgery? This would allow us save both dollars and disutility for patients.}'' The principles from Section \ref{sec:interventions} show that this example fails to describe the intervention through which prediction would actually save dollars and disutility. Would predicting which surgeries be futile really be the best target here given who will be seeing the predictions? Showing this to doctors could have further externalities of either making doctors overconfident or reducing patient choice (similar to observations in Section \ref{sec:interventions:harms}). Are the inputs mutable or conclusions actionable for the patient (Section \ref{sec:interventions:actionability})? Through these observations and questions, our education study foregrounds a key piece of the translation between predictions and interventions which has previously been taken for granted.

Emerging work in causal inference suggests ways to develop intervention models from data as opposed to prediction models in various domains \citep[e.g.][]{motz2018embedding,prosperi2020causal}, but existing causal inference approaches such as randomized controlled trials have known challenges with external validity \citep{Stephenson98RCT,deaton2018understanding}; as such, identifying actionable insights with causal methods remains an ongoing project. Moreover, Kohler et al. \cite{kohler2018eddie} and Hu et al. \cite{hu2020sex} have pointed out significant conceptual flaws with interpreting social categories such as race and gender as causally manipulable variables (such as in a causal diagram \citep{pearl1995causal}), suggesting that the validity of causal inference cannot be taken for granted where demographic aspects of student data are concerned. 

%The above references discuss inconsistencies with the counterfactual thinking behind “what if the individual belonged to a different demographic group”. It is problematic to treat race or gender or any social category as manipulable via a do-operation. To make such an operation valid (i.e. valid to consider the prediction outcome of ‘flipping' a person's race) requires extremely unreasonable assumptions on the world (e.g. that such a person can exist in the our version of the world, that race is not inherently entangled with any number of other socioeconomic features that remain unchanged)

%\lnote{Actionability implies causality but not necessarily other direction. Actionability means the change must be what you want, i.e. improvement in the variable of interest, rather than a proxy.}

\subsection{Extensions and future work}
As our study sheds light on gaps in problem formulation and interventions in ML research, we also highlight a number of open questions in how to further bridge these gaps. For example, given the lack of a clear single objective in education problems shown in Section \ref{sec:problem_formulation}, the formulation of ML problems to navigate the ecosystem of multiple education goals and stakeholders is still a significant open problem. Also, more concrete exploration of how to produce actionable insights using ML would be an impactful avenue of algorithmic study.

There are several extensions to our methodology which could lead to further insights in education and beyond. Interviews with education researchers produce a unique set of discussions which is distinct from interviews with students or teachers. Thus, further fieldwork with direct stakeholders would provide a different and valuable dimension of insight. It may also be illuminating to supplement the qualitative interview study with quantitative statistics from a larger set of applied ML papers. For other domains such as healthcare and criminal justice, this methodology may be applied directly to examine ML research papers and perhaps uncover subtle tensions that were less prevalent in education.

Finally, in the critical evaluation of research practices, another avenue to explore is the degree of cross disciplinary collaboration in applied ML research papers, and relatedly, a comparison of power structures and research incentives in the different fields. Our interviews hinted at the two research fields placing different value judgements for different types of problem formulation, with education researchers trending towards causal interventions and ML researchers preferring predictive modeling. This opens the door for a future exploration of the influence of these competing forces in applied ML research collaborations.

\paragraph{Acknowledgements} We are deeply grateful to our interview participants for sharing their invaluable time and insights with us. We thank Gloria Chua for their indispensable help with graphic design (Figure 1). We thank anonymous reviewers, Solon Barocas, Marzyeh Ghassemi, Jon Kleinberg, Karen Levy, and Sam Robertson for their vital and constructive feedback on earlier iterations of the work. Last but not least, we thank our colleagues at various seminar talks for their thought-provoking questions and comments.

% Bibliography
\bibliographystyle{abbrvnat}
\bibliography{ml4ed}

\end{document}